\documentclass[runningheads]{llncs}
\usepackage[T1]{fontenc}
\usepackage{graphicx}
\usepackage{algpseudocode}
\usepackage{tablefootnote}
\usepackage{makecell}
\usepackage{booktabs}
\usepackage{amsmath}
\usepackage{amsfonts}

\begin{document}
\title{Neural Contrast: Leveraging Generative Editing for Graphic Design Recommendations}
\titlerunning{Neural Contrast}
% If the paper title is too long for the running head, you can set
% an abbreviated paper title here
%\

\author{Marian Lupa\textcommabelow{s}cu\inst{1,2} \and
Ionu\textcommabelow{t} Mironic\u{a}\inst{1} \and
Mihai Sorin Stupariu\inst{2}}
%\author{ANONYMOUS AUTHOR(S)}

\authorrunning{M. Lupa\textcommabelow{s}cu et al.}
%\authorrunning{Anon. Submission}

% First names are abbreviated in the running head.
% If there are more than two authors, 'et al.' is used.
%
\institute{Adobe Research \and University of Bucharest, Romania}

\maketitle              % typeset the header of the contribution
\begin{abstract}
Creating visually appealing composites requires optimizing both text and background for compatibility. Previous methods have focused on simple design strategies, such as changing text color or adding background shapes for contrast. These approaches are often destructive, altering text color or partially obstructing the background image. Another method involves placing design elements in non-salient and contrasting regions, but this isn't always effective, especially with patterned backgrounds. To address these challenges, we propose a generative approach using a diffusion model. This method ensures the altered regions beneath design assets exhibit low saliency while enhancing contrast, thereby improving the visibility of the design asset.

\end{abstract}
\section{Introduction}

\begin{figure*}[h]
  \centering
  \includegraphics[width=\linewidth]{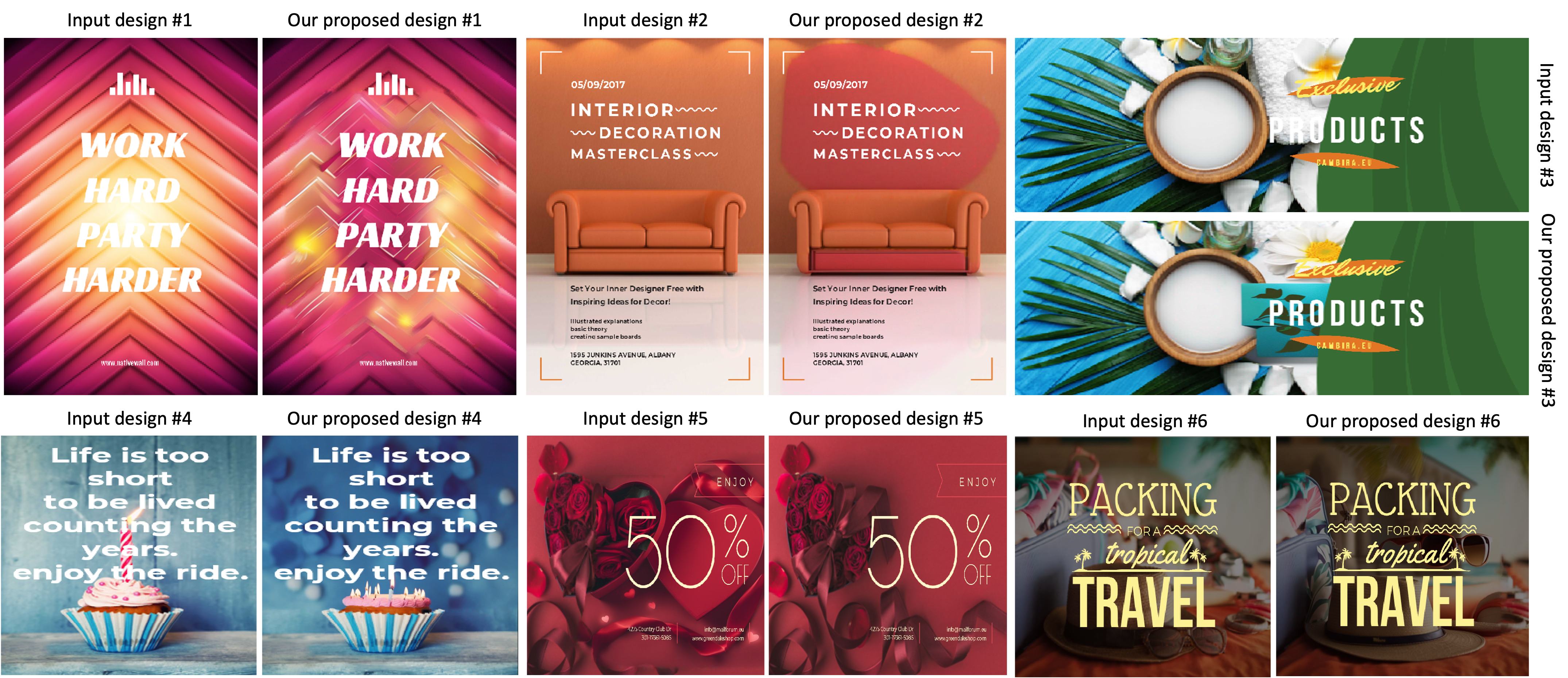}
  \caption{\textmd{Our generative editing pipeline enhances contrast in graphic design elements, such as text or SVGs, on background images. This pipeline automatically adjusts the luminance to contrast with text color, thereby emphasizing the text. As shown in examples 1, 4, and 6, luminance changes in the opposite direction of text color to enhance contrast. Examples 2 and 3 demonstrate text emphasis through the creation of contrasting elements. Additionally, contrast is further improved by reducing image energy, such as removing background objects behind the text, as seen in examples 4 and 5. This automated process involves injecting color, luminance, content, and fractal noise, followed by generative editing with a diffusion model.}}
  \label{fig:teaser}
\end{figure*}

Compositing, the process of integrating distinct visual elements into a cohesive design, is a fundamental aspect of graphic design. Achieving visual authenticity and appeal in these composites requires careful consideration of the compatibility between text and background. However, previous efforts to harmonize these elements have faced limitations in effectiveness, often focusing predominantly on text color preservation while neglecting additional design elements such as background shapes and the systematic regeneration of image backgrounds for optimal contrast. These shortcomings are particularly problematic when text occupies a substantial portion of the design, as enhancing contrast and legibility can significantly alter the overall layout and confuse users \cite{meron2022graphic,huang2023survey}.

Our proposed solution addresses these limitations by implementing a nuanced strategy that incorporates improvements in text placement and background modifications to emphasize the text. These detailed adjustments are systematically provided to users as a catalog of visual variations, enabling quick and efficient selection of the most appropriate option without extensive manual intervention. This approach ensures a cohesive integration of text and background elements, preserving overall design integrity and enhancing user experience.

\section{Related Work}

Current state-of-the-art methods for improving graphic designs focus on three main approaches: color recommendation, layout recommendation, and text integration.

Color recommendation methods can be divided into photo color adjustment and design element recoloring.

\textbf{Photo Color Adjustment \cite{1zhao2021selective,2qiu2023color}}: These methods involve editing the colors of images within a design. Partial editing changes the color of specific objects that deviate from the design's palette, while total editing involves color transfer. \cite{1zhao2021selective} uses a region selection network and a recoloring network for partial editing, while \cite{2qiu2023color} performs complete recoloring of background images using a color palette derived from a BERT model.

\textbf{Recoloring Design Elements \cite{4hegemann2023cocolor,qiu2023multimodal,3shi2023stijl}}: These methods recolor easily adjustable design elements, using colors extracted from prominent regions within the existing design. \cite{4hegemann2023cocolor} uses a color extraction module and an image-to-image color recommendation model. \cite{qiu2023multimodal} employs a multimodal masked color model incorporating CLIP for text embeddings to recommend colors based on both color and textual contexts. \cite{3shi2023stijl} introduces De-Stijl, a system with a 2D palette extractor, color recommender, and image recolorizer for creating harmonious color graphics.

\begin{figure*}[h]
  \centering
  \includegraphics[width=\linewidth]{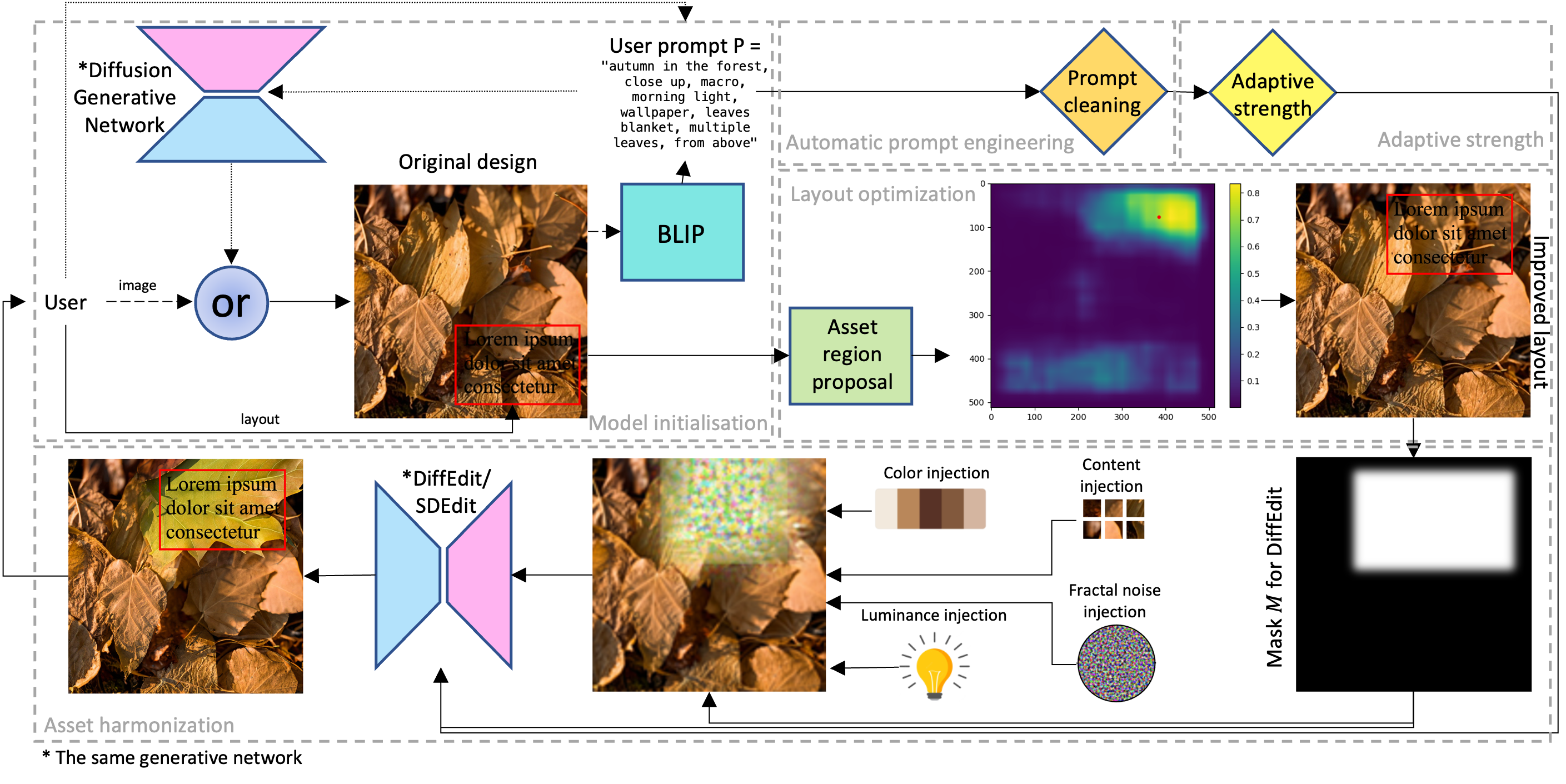}
  \caption{\textmd{An overview of the design editing/generative pipeline, detailing all intermediary stages and outcomes, except for Design Variations. The input prompt "autumn in the forest, close up, macro, morning light, wallpaper, leaves blanket, multiple leaves" forms the input of the pipeline. Dashed lines separate distinct steps, with Design Variation as the exception. Diamond symbols denote algorithms, while rectangular shapes represent models. The pipeline input comprises the layout combined with either the $P$ prompt alone, the design background alone, or both.}}
  \label{fig:pipe}
\end{figure*}

Layout recommendation methods fall into two categories: those based on design principles and those using generative models.

\textbf{Design Principles \cite{5o2014learning}}: These methods propose layout suggestions based on principles like spacing, contrast, similarity, and white space, ranked by a model trained on desirable design patterns.

\textbf{Generative Models \cite{6inoue2023layoutdm,7li2020attribute,8yamaguchi2021canvasvae,hsu2023posterlayout,lin2023autoposter}}: These methods train generative models (e.g., GANs, VAEs, diffusion models) to create layouts conditioned on existing designs or constraints. \cite{6inoue2023layoutdm} uses discrete diffusion for variable-length layouts, \cite{7li2020attribute} integrates attributes like area and aspect ratio, and \cite{8yamaguchi2021canvasvae} employs a VAE for document reconstruction and generation. \cite{hsu2023posterlayout} and \cite{lin2023autoposter} focus on content-aware visual-textual layouts for posters and advertisements.

A \textbf{text integration technique} is TextDiffuser \cite{chen2024textdiffuser} and introduces a method for generating visually appealing text images using a combination of Transformer and diffusion models. It involves layout generation with a Transformer to locate keywords and generate segmentation masks, followed by image generation with a diffusion model. 

\section{Approach}

The method comprises five distinct steps, delineated by dashed lines in Figure \ref{fig:pipe}. Each step is detailed in the subsequent subsections. Additionally, an optional Design Variation step can be performed, involving the execution of the five steps of the pipeline on the same text and image, with optimal variation of text properties.

\subsection{Model initialisation}

In the absence of a user-provided background image, it is necessary to obtain a prompt $P$ from the user to serve as the initial input. This prompt is crucial because a design requires a foundational concept, whether in the form of text or an image, to describe the intended background image. Leveraging the sophisticated image generation capabilities of diffusion networks in contemporary generative tasks \cite{croitoru2023diffusion,dhariwal2021diffusion}, we employ a generative diffusion model conditioned on the user-provided prompt $P$ to generate the required background image.

When the user provides an background image, background image generation is not required. If the user opts not to overlay a design asset on a background image, the default background is set to white. In cases where a design asset is to be overlaid, the problem is segmented into subproblems involving the design asset with either a background or a white image. If no prompt is available for conditioning the diffusion model, generation is initiated based on the input image. This is accomplished by utilizing the BLIP2 captioning model \cite{11li2023blip} to extract the prompt $P$ from the provided image. 

\subsection{Automatic prompt engineering} \label{IP}

The user-provided prompt $P$ often contains chromatic details that hinder the denoising process from producing a more legible design asset. A higher presence of chromatic information in $P$ makes it more challenging for the diffusion model to improve legibility through contrast and emphasize the desired design asset.

To address this, we perform a prompt cleaning preprocessing step, replacing all chromatic information in $P$ with a color opposite to that of the design asset. This approach balances the chromatic vector direction in the denoise process, optimizing the diffusion process by either reducing the number of denoise steps required or improving quality given a constant number of steps. The opposite color is calculated through a three-dimensional binary search within the RGB space using the $CIELAB \Delta E*$ distance in the 2000 formulation. 

\subsection{Layout Optimization} \label{LO}

The SAM model \cite{cornia2018predicting}, which integrates an Attentive Convolutional Long Short-Term Memory network \cite{sainath2015convolutional}, is highly effective at estimating the visual strength of design templates. It captures human attention priors, such as the center bias, making it ideal for scenarios where certain design elements need more focus. This is why we chose SAM to approximate a salience map for design templates.

Let $H$ be defined as $SAM(T)$, where $T$ is the input design template, and $SAM(\cdot)$ is the SAM model. The complementary counterpart, $1-H$, functions as a heat map, highlighting regions with low saliency. To optimize the layout, design assets are strategically placed to maximize their dispersion within $1-H$, avoiding overlap with other elements. This ensures an effective and visually appealing arrangement by positioning assets in non-salient areas to enhance their emphasis.

\subsection{Adaptive strength} \label{AS}

\begin{figure*}[h]
  \centering
  \hspace*{-2cm}
  \includegraphics[scale=0.6]{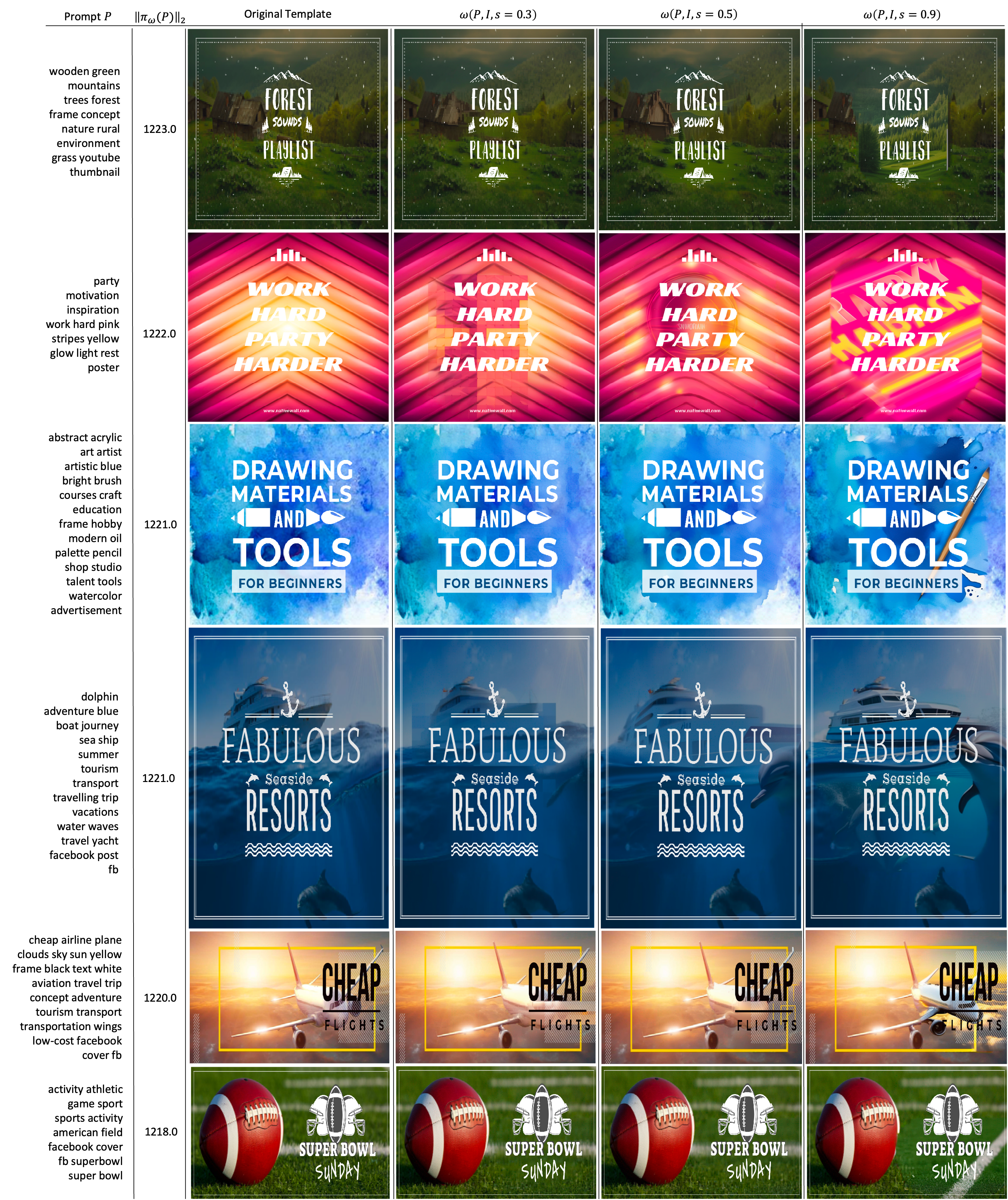}
  \caption{\textmd{The investigation focuses on evaluating the influence of the strength parameter on the SDXL-base model \cite{podell2023sdxl} within a subset of the utilized data sourced from Crello \cite{8yamaguchi2021canvasvae}. The objective is to assess the impact of the strength parameter on the final outcomes of each model, where $\omega\left(\cdot, \cdot, \cdot\right)$ is the Diffusion Generative Network in the DiffEdit \cite{13couairon2022diffedit} or SDEdit paradigm \cite{14meng2021sdedit}, where the first parameter is the constant prompt $P$ for all generated images, the second parameter is the constant starting image $I$ (obtained precisely as in Subsection \ref{AH}) for all generated images and the third parameter is the intensity that is desired from $I$, or the strength parameter from SDEdit and DiffEdit.}}
  \label{fig:AS}
\end{figure*}

In our preliminary experiments using the pipeline depicted in Figure \ref{fig:pipe}, excluding the Adaptive Strength feature and maintaining a constant strength parameter in the diffusion models \cite{ho2022classifier} across different seeds, we observed an inconsistency. Despite the consistent strength parameter, the generative model exhibited varying degrees of aggressiveness (output input difference) with different prompts. Particularly intriguing was the consistent level of aggressiveness observed across different seed configurations using identical prompts. This inconsistency prompted further investigation into its underlying cause. Subsequently, we identified that the norm of the projection of the prompt in the text embedding space was responsible for this behavior, as illustrated in Figure \ref{fig:AS}.

For prompts with high norm values, lower strength initially produced more natural and harmonious outputs, while higher strength caused more aggressive and structurally modified outputs. Conversely, prompts with low norm values displayed the opposite behavior, with lower strength resulting in altered outputs and higher strength yielding more natural outputs (as illustrated in the last row of Figure \ref{fig:AS}, where the white line of lime behind the text is altered at strength 0.9). This variation arises from the changing distance between the prompt projection and the center of the hypersphere in the text embedding space. To resolve this, we approximated the strength based on the norm of the prompt projection in the latent space. This strength parameter is then used in the generative model $\omega$ within both the SDEdit and DiffEdit paradigms, defined as follows:

\begin{equation}
\label{eq:stre}
  strength=\ \mathcal{F}(\| \pi_{\omega}(P) \|_2 ),
\end{equation}

\begingroup
\raggedright 
where $\mathcal{F}:\mathbb{R}\rightarrow[0,1]$ is a function that approximates the strength for any norm of a prompt $P$ projected in the text embedding space of the diffusion model, with projector $\pi_{\omega}$, and $\| \cdot \|_2 $ is the $L_2$ norm. 
\endgroup

\begin{figure}[h]
  \centering
  \includegraphics[width=\linewidth]{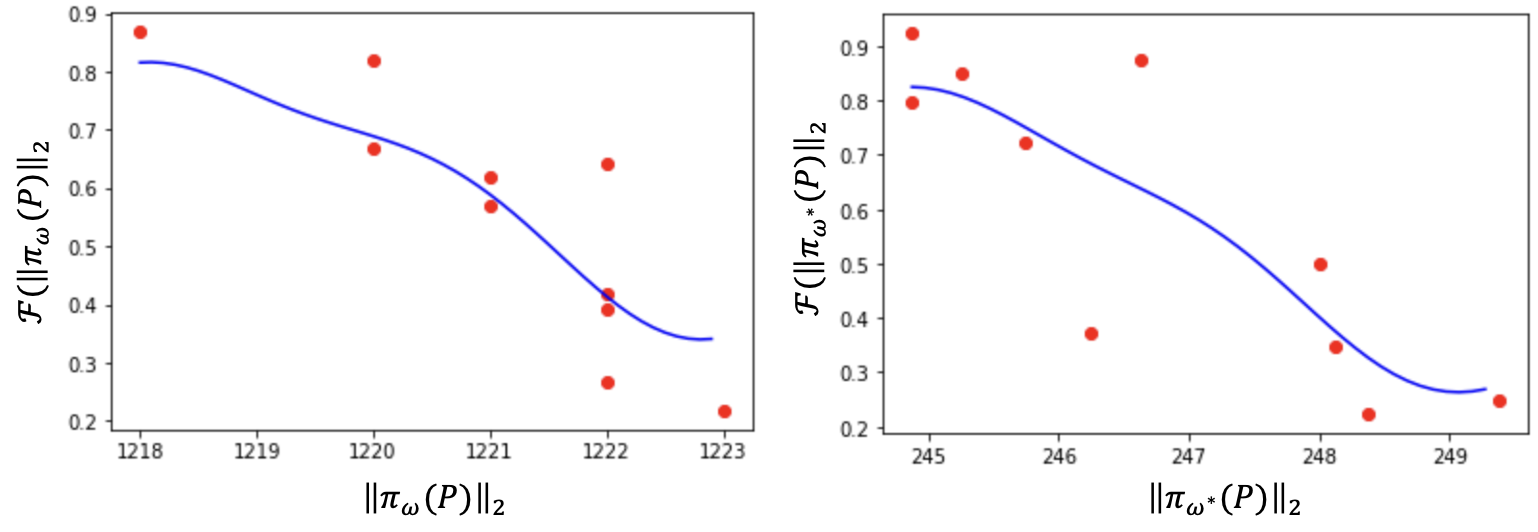}
  \caption{\textmd{The plot of the function $\mathcal{F}$ post-training using the dataset pairs $X$ for input data and $Y$ for corresponding labels. Left: $\mathcal{F}$ function for the model $\omega=$ SDXL-base \cite{podell2023sdxl} model. Right for the model $\omega^*=$ SD-v1.5 \cite{rombach2022high}}}
  \label{fig:ASpattern}
\end{figure}

We chose $\mathcal{F}$ to be a Support Vector Regression (SVR)\cite{drucker1996support} because: SVR excels in high-dimensional spaces, is robust to overfitting, and handles non-linear relationships. It is effective at capturing complex data patterns, robust to outliers, and offers good generalization. As shown in Figure \ref{fig:ASpattern}, where there is only two outliers (for SD-v1.5). We select RBF (a Gaussian type) kernel for more non-linearity.

We conduct training the SVR on a dataset comprising $N$ data points. The data collection process involves generating $k \cdot N$ input images corresponding to $N$ prompts and  $k > 1$ seeds. For each prompt, we calculate the projection norm in the text embedding latent space (these norms represent $X$ set). Then for these $k \cdot N$ images diffusion process was applied with strength $s \in S \subseteq [0,1]$. Then the best strengths for each prompt were manually selected (the best strengths represent $Y$ set). Subsequently, the training of $\mathcal{F}$ was conducted utilizing a dataset denoted as $X$ along with corresponding labels noted $Y$.

 In our particular scenario, we set $N$ to $10$, $k$ to $2$, and $S$ to the discrete sequence $\left\{0, 0.1, 0.2, \ldots, 0.9\right\}$. The choice of only 10 prompts and two seeds was deliberate, given that Support Vector Regressors exhibit ease of learning and do not necessitate an extensive amount of training data, as highlighted in previous research \cite{drucker1996support}. Additionally, the discernible pattern within the data further justifies this limited selection.

\vspace{-1cm} \subsection{Asset harmonization} \label{AH}

This process involves editing the original image to enhance the readability and harmony of overlapping design assets. The image alteration occurs in two main phases. In phase one, an auxiliary image is created by injecting contrast from three sources into the original image. In phase two, the auxiliary image is iteratively modified using a diffusion process based on a modified prompt and automatic strength. These changes are confined to the intersection of the image and the design asset, following the DiffEdit paradigm \cite{13couairon2022diffedit}, using a Gaussian mask proportional to the design asset's size. Alternatively, in the SDEdit paradigm \cite{14meng2021sdedit}, the entire image is modified. The paradigm is user-selected.

In phase one contrast is injected from three sources: luminance, opposite color, and fractal noise. Luminance visibly increases readability because it is directly related to the contrast ratio calculation \cite{ma2023first}. Moreover, an increase in luminance in any direction also represents a decrease in the salience of the modified area. This is beneficial for the readability and harmony of the design asset over the overlay image. Luminance is injected according to the following formula:

\vspace{-0.4cm}\begin{equation}
\label{eq:lum}
\begin{split}
    \Delta L = sgn\left(L\left(A\right)-0.5\right) \cdot \frac{1}{\alpha-\beta\left|L\left(A\right)-L\left(B\right)\right|}, \\ 
    \end{split}
\end{equation}

\begingroup
\raggedright 
where $A$ represents the color of the design asset in the RGB color space, $B$ represents the average color of the image portion with which the design asset overlaps in the RGB color space, $L\left(K\right)$ represents the luminance calculation function for a color $K$, $sgn\left(x\right)=\frac{\left|x\right|}{x}$ is the sign function. The calibration parameters, $\alpha=\frac{1}{min}$ and $\beta=\alpha-\frac{1}{max}$, are indicative of the luminance injection intensity. Here, $\Delta L \in[-max,\ -min] \cup [min,max]$, with the calibrated interval determined through multiple iterations, resulting in $min=0.2$ and $max=0.8$ in our experiments. The variable $\Delta L$ signifies the luminance injection amount into the region of interest of the design asset. The greater the luminance difference between the two entities, the less the need to change the luminance, and vice versa.
\endgroup

Enhancing the visibility of overlapped assets in the SDEdit or DiffEdit edited region is achieved through contrasting color injection. This method computes the complementary color (as described in Subsection \ref{IP}), employs a local point coloring model guided by this color, and generates local points using a semantic segmentation model \cite{16wang2023one}. Only objects with at least 20\% mask coverage and 80\% confidence are considered. Furthermore, the entire corrected mask region is colored with the complementary shade to promote consistency and reduce prominence. The intensity of the added color is inversely proportional to the existing color in the mask region, defined by the formula:

\vspace{-0.3cm}
\begin{equation}
\label{eq:color}
\begin{split}
    \Delta H = \Phi\left(M\middle|\begin{matrix}\Gamma\left(M,\ O\right)\geq0.8\ \\\mathcal{A}\left(O\right)\geq0.2\mathcal{A}\left(M\right)\\\end{matrix}\right) \cdot \frac{M\cdot C}{\alpha-\beta \cdot \widetilde{\sum_{\substack{\left(i,\ j\right)\in M \\ \pi_{HSV\ }\left(M\left[i,j\right]\right)\ \in\mathcal{V}\left(C\right)}}1}},
\end{split}
\end{equation}

\begingroup
\raggedright 
where $\Phi(\cdot|\cdot)$ denotes the recoloring model with the first parameter being the recoloring image (here, mask $M$) and the second parameter representing the optimal condition set of local points, $\Gamma\left(M,\ O\right)$ is the semantic segmentation model score for image $M$ and identified object $O$, $\mathcal{A}\left(\cdot\right)$ is the area function, $\pi_{HSV\ }\left(\cdot\right)$ projects from RGB to HSV space, $\mathcal{V}\left(\cdot\right)$ defines the color neighborhood using the inRange function, and $\widetilde{\ \ \cdot\ \ }$ normalizes a vector (specifically $\widetilde{v}=\frac{v}{\max(| v |_2, \epsilon)}$ with $\epsilon=10^{-12}$). The constants $\alpha$ and $\beta$ are as defined in Equation \eqref{eq:color}. Although multiple contrasting colors can theoretically be employed, our experiments predominantly utilize a single color due to the predominantly monochromatic nature of design assets.
\endgroup

Furthermore, maintaining consistent color palette integrity across content is pivotal in achieving uniformity in the final output of a scene. This is achieved by injecting and extracting content from the original template. Specifically, we identify $N$ patches of size $B \times B$ in the background image of the design template, adhering to criteria where each patch's average color contrasts by at least $4.5$ with the design asset collection \cite{ma2023first}, and the patch's pixel standard deviation is no less than $0.05$. These patches, termed diffusion-friendly, are less prone to distortion by diffusion-based generative models. Once identified, these patches are integrated into a texture aligned with the design asset region dimensions using methods from \cite{efros2023image}, with parameters $N=1000$ and $B=25$. Additionally, fractal noise, detailed in \cite{17lupascu2022fast}, is introduced to indirectly enhance contrast through observed forgetting phenomena. Diffusion models creatively enhance areas with injected noise, accentuating contrast injections and diminishing the impact of existing content.

In the subsequent phase, a new auxiliary content image is obtained, either entirely modified by SDEdit or partially adjusted with DiffEdit around the mask derived from the diffusion model’s denoising process. This denoising process follows prompt conditions as in Subsection \ref{IP} and is calibrated to strength settings as in Subsection \ref{AS}. Ultimately, the resulting image prominently features the superimposed design asset, ensuring readability and visual prominence.

\subsection{Design variation} \label{DV}

Design Variation involves creating several outputs from a single set of user inputs. In our experiments, we set the number of variations to 4. The user’s inputs lead to two cases for generating variations: with an existing design or without one. If a design from the user is given, then the first two variations are generated from the user's design (that is, all steps are executed apart from Subsection \ref{LO}) and the other two are obtained using Subsection \ref{LO}. 

Figure \ref{fig:DV} illustrates several instances. The initial two variations align with Subsection \ref{AH}, employing identical auxiliary images, executed only once due to identical inputs. The subsequent two variations, following Subsection \ref{LO}, mirror the approach of the initial pair. Variations two and four share colors randomly selected from the design template palette. If the user does not provide any layout, then all four variations are obtained exactly like the last two variations above.

\section{Results and Experiments}

We demonstrate the versatility and adaptability of our generative editing approach, highlighting its effectiveness across diverse design scenarios that require emphasizing specific design assets. Our evaluation includes qualitative assessments using the complete pipeline shown in Figure \ref{fig:pipe} and quantitative analyses by disabling design variations and asset region proposals for direct comparison with established methods in the literature. We used the models SDXL-base \cite{podell2023sdxl} and SD-v1.5 \cite{rombach2022high} for all experiments, though our method is general and can utilize any generative model that accepts the strength parameter. The only requirement is recalculating the function $\mathcal{F}$, where the adaptive strength depends on the generative model $\omega$.

\subsection{Test Set}

We evaluate the quantitative performance of generative editing using the Crello \cite{8yamaguchi2021canvasvae} test set, which underwent meticulous manual refinement. This refinement process involved eliminating templates exhibiting rendering errors and excluding those templates that do not embody a valid use case for generative editing. Specifically, templates featuring design assets superimposed on monochrome surfaces (such as SVGs or backing shapes) were excluded, as manipulating the monochrome area lacks meaningful application. Subsequently, a secondary filtering was conducted to exclusively retain design assets containing at least one textual component. Moreover, the examples used to calculate the function $\mathcal{F}$ were excluded. After performing the above mentioned filtering steps, 587 valid templates remained. All the methods used in Table \ref{tab:metrics} were evaluated on this subset of the Crello test set. In our experiments, all existing texts in the designs were considered as design assets. The prompts used are given by a concatenation of all keywords for the design templates from Crello.

\subsection{ Comparison to Prior Methods}

We compare our method with the following methods:
\begin{itemize}
  \item Selective Region-based Photo Color Adjustment for Graphic Designs \cite{1zhao2021selective} explore region-based photo color adjustment in graphic design. Their model includes a region selection network trained on natural images and design knowledge, along with a recoloring network, addressing challenges in locality and naturalness.

  \item FlexDM \cite{6inoue2023layoutdm} trained a holistic model for multi-modal document design. Treating vector graphic documents as collections of elements, it utilizes a unified architecture to predict masked fields. For the purpose of our study, we specifically concentrate on the image task ([img]) due to its relevance for a direct comparison with our specific use case.

\item TextDiffuser \cite{chen2024textdiffuser} is a two-stage framework for generating text-integrated images using diffusion models. Our focus is on the inpainting task, relevant for enhancing text prominence. Unlike other methods, which compare original and stylized results directly, for TextDiffuser, our evaluation uses an inpainting scenario with 5 seeds and the prompt "Design of [words used in the design rendering]," keeping other parameters default. Scores are averaged across the original and TextDiffuser-generated images for all five seeds.

\end{itemize}

\subsection{Quantitative Results}

\begin{table*}
  \caption{\textmd{Scores of different methods on Crello test set}}
  \label{tab:metrics}  
  \begin{tabular*}{\linewidth}{lcccccccccc}
    \toprule
    Methods & PNSR $\uparrow$ & SSIM $\uparrow$ & LPIPS $\downarrow$ & SDI $\downarrow$ & $R_{com}$ $\downarrow$ & SAM $\downarrow$ & FID $\downarrow$ & VIF $\uparrow$ & AS $\uparrow$ \\
    \midrule
    Original test set & 100 & 1 & 0 & 0 & 14.60 & 0 & 0 & 1 & 3.89  \\
    Zhao et al. 2021 \tablefootnote{\label{first}Findings derived from the outcomes presented in the paper} & 31.25 & 0.78 & 0.19 & 0.98 & 14.60 & 0.11 & 69.59 & 0.63 & 3.90  \\
    FlexDM [img] & 32.02 & 0.71 & 0.21 &  0.40 &14.60 & 0.15 & 65.23 & 0.69 & 3.59  \\

    \makecell[l]{TextDiffuser \\ $[$inp$]$ [SD-v1.5]} & 31.48 & 0.70 & 0.29 &  0.28 & 15.05& 0.08 & 77.88 & 0.48 & 3.50  \\

    \makecell[l]{TextDiffuser \\ $[$inp$]$ [SD-v2.1]} & 31.18 & 0.69 & 0.31 & 0.32 &15.10 & 0.09 & 95.66 & 0.44 & 3.66 \\

    \makecell[l]{Our - adaptive \\ strengt [SD-v1.5]} & 32.27 & 0.76 & 0.21 & 0.22 &14.42 & 0.03 & 40.76 & 0.79 & 3.76 \\
    
    \makecell[l]{Our + adaptive \\ strength [SD-v1.5]} & 32.76 & 0.78 & 0.19 &  0.22 & \textbf{14.36} & \textbf{0.03} & \textbf{36.09} & 0.82  & 3.90  \\
    
    \makecell[l]{Our - adaptive \\ strengt [SDXL-b]} & 34.12 & 0.86 & 0.19 & 0.20 &14.94 & 0.04 & 56.12 & 0.90 & \textbf{3.95} \\
    \makecell[l]{Our + adaptive \\ strength [SDXL-b]} & \textbf{34.46} & \textbf{0.87} & \textbf{0.17} & \textbf{0.20} &14.56 & 0.03 & 50.43 & \textbf{0.92} & 3.95\\
  \bottomrule
\end{tabular*}
\end{table*}

Initial computations involve the Peak Signal-to-Noise Ratio (PSNR) assessment across various methods applied to our test dataset, detailed in Table \ref{tab:metrics}. Our proposed methodology achieves the highest PSNR score among previous approaches, demonstrating the efficacy of our editing technique. Adaptive strength enhances results by aligning the editing pipeline's intensity with the contrast and color variation between the emphasized design asset and its background.

Structural Similarity Index Measure (SSIM) \cite{nilsson2020understanding}, Learned Perceptual Image Patch Similarity (LPIPS) \cite{zhang2018perceptual} and Spatial Distortion Index (SDI) \cite{Alparone2008MultispectralAP} gauge structural fidelity and perceptual similarity compared to the original image. Our method utilizing adaptive strength with the SDXL-base ([SDXL-b] in Table \ref{tab:metrics}) model consistently outperforms others in these evaluations. 

In design contexts, strategic text placement on uncluttered, visually prominent areas ensures readability and balance. Readability and Visual Balance ($R_{com}$) \cite{zhou2022composition} quantifies layout quality, revealing that our approach with SD-v1.5 tends to remove objects behind text, while SD-v2.1 enhances contrast by introducing new edges and objects around text.

Spectral Angle Mapper (SAM) optimizes image composites by enhancing spectral signatures, improving contrast and visual compatibility. Our approach consistently outperforms existing methods across all models assessed in the literature. Frechet Inception Distance (FID) \cite{heusel2017gans} quantitatively assesses the quality of generated images, crucial for evaluating visual appeal and text-background compatibility in generative methodologies. Notably, our approach achieves optimal FID results with the SD-v1.5 model, suggesting nuanced trade-offs in image naturalness compared to stronger models.

Additionally, we evaluate the aesthetic quality of our results in comparison with \cite{1zhao2021selective} and \cite{6inoue2023layoutdm} using the VIT-L-14 aesthetic predictor from LAION-AI \cite{kim2023architectural}, presented in Table \ref{tab:metrics}. Our method achieves superior aesthetic outcomes without employing adaptive strength. The Visual Information Fidelity (VIF) \cite{1576816}, aligned with human visual perception, also highlights our method's performance, particularly when employing adaptive strength with the SDXL-base model.

%\begin{figure}[h]
%  \centering
%  \includegraphics[width=\linewidth]{fail.png}
%  \caption{\textmd{Failure cases. The augmented emphasis on design output templates within the text area has led to an elevated contrast ratio between text and background. Despite this increase, the overall readability, as well as the naturalness and harmony of the design, has diminished.}}
%  \Description{Failure cases}
%  \label{fig:fail}
%\end{figure}

\subsection{Qualitative Results}

\begin{figure*}
  \centering

  \vspace{-0.2cm}
  \hspace*{-1.7cm}\includegraphics[width=14cm]{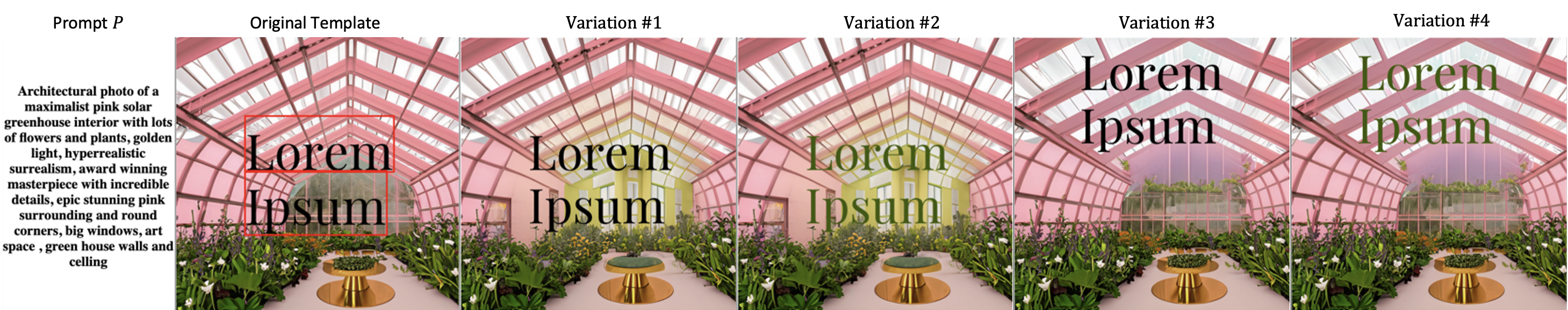}
  \hspace*{-1.7cm}\includegraphics[width=14cm]{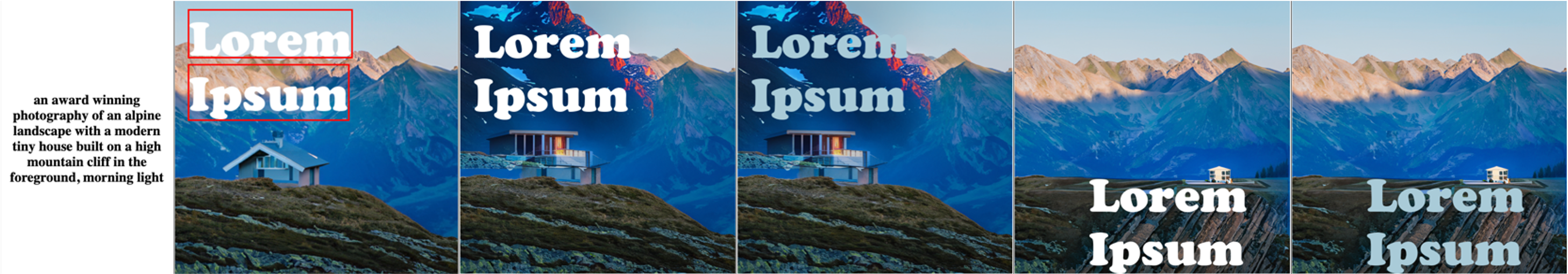}
  \hspace*{-1.7cm}\includegraphics[width=14cm]{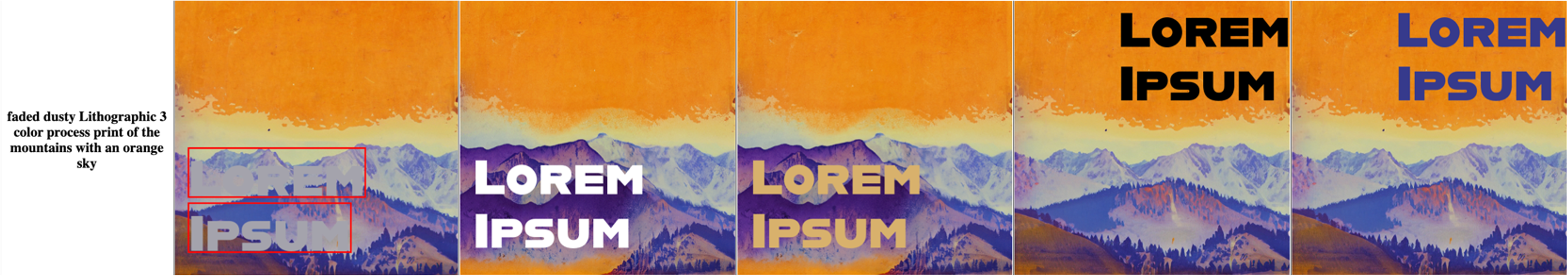}
  \hspace*{-1.7cm}\includegraphics[width=14cm]{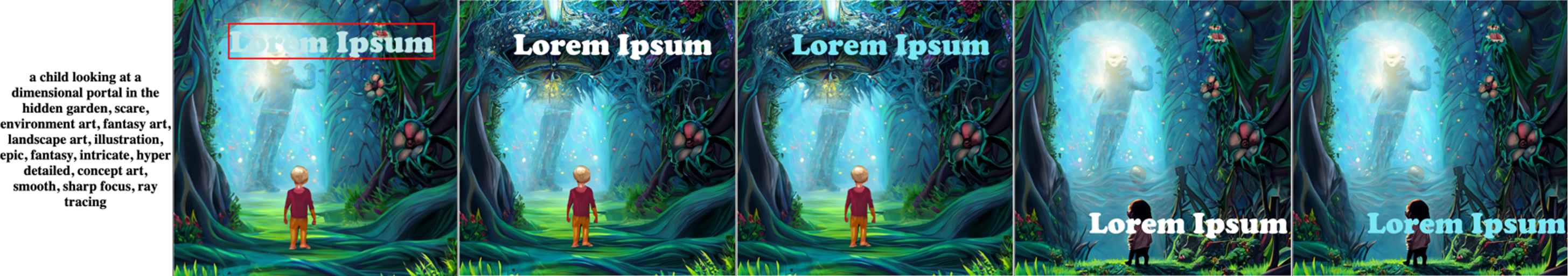}
  \hspace*{-1.7cm}\includegraphics[width=14cm]{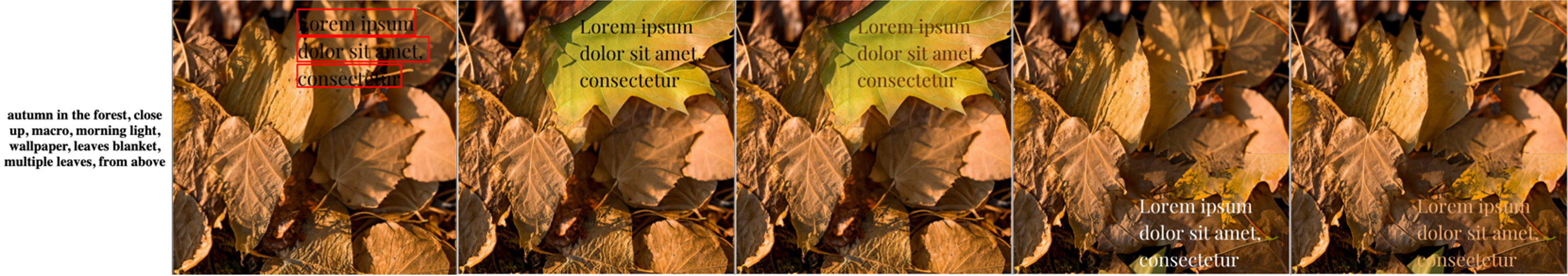}
  %\hspace*{-2cm}\includegraphics[width=14cm]{Picture_6.png}
  \caption{Design Variation Potential: The first column shows the input prompt. The second column displays the initial design, marked with a red border for visibility at reduced readability. The next two columns present variations of the original layout, followed by the final two columns which feature new layout-generated variations.}
  \label{fig:DV}
\end{figure*}

Figure \ref{fig:DV} illustrates the qualitative outcomes achieved by our comprehensive pipeline (refer to Subsection \ref{fig:DV} for quantitative results). Our approach systematically enhances the original design, leading to a seamless improvement in text visibility. In Figure \ref{fig:DV}, there are instances where readability poses challenges, but our proposed method is capable of addressing such cases. In contrast to conventional methods that concentrate on text or background color, our approach proves to be more effective in tackling this issue.

\section{Conclusions and future work}
\vspace{-0.2cm}
Our methodology utilizes generative editing techniques to improve graphic design quality by overcoming traditional limitations in handling text and background elements. Neural Contrast employs a diffusion model to adjust underlying regions beneath design assets, reducing saliency in specific areas and enhancing contrast to emphasize design elements. Our contributions include a comprehensive pipeline for modifying graphic designs, demonstrating the potential of generative editing to optimize layouts and highlight design assets. This approach diverges from conventional methods by enhancing contrast and legibility while structurally modifying design templates. Experimental results showcase its versatility across diverse design scenarios, suggesting its ability to advance graphic design through integration of generative editing.

%
% ---- Bibliography ----
%
% BibTeX users should specify bibliography style 'splncs04'.
% References will then be sorted and formatted in the correct style.
%
% \bibliographystyle{splncs04}
% \bibliography{mybibliography}
%

\bibliographystyle{splncs04}
\vspace{-0.3cm}
\bibliography{samplepaper}

\section{Appendix}

\end{document}